\documentclass[runningheads,a4paper]{llncs}

\usepackage{amssymb}
\usepackage{graphicx}
\usepackage{color}
\usepackage{multirow}

\usepackage{amsmath}        
\usepackage{bm}             
\usepackage{graphics}       
\usepackage{etoolbox}       

\usepackage{array}
\newcolumntype{L}[1]{>{\raggedright\let\newline\\\arraybackslash\hspace{0pt}}m{#1}}
\newcolumntype{C}[1]{>{\centering\let\newline\\\arraybackslash\hspace{0pt}}m{#1}}
\newcolumntype{R}[1]{>{\raggedleft\let\newline\\\arraybackslash\hspace{0pt}}m{#1}}

\begin{document}
	
	\mainmatter  
	
	\title{End-to-End Unsupervised Deformable Image Registration with a Convolutional Neural Network}
	
	\titlerunning{Deformable Image Registration ConvNet}
	
	%
	%
	
	\author{Bob D. de Vos\textsuperscript{1}
		\and
		Floris F. Berendsen\textsuperscript{2}
		\and
		Max A. Viergever\textsuperscript{1}
		\and
		Marius Staring\textsuperscript{2}
		\and\\
		Ivana I\v{s}gum\textsuperscript{1}}
	
	\authorrunning{Bob D. de Vos et al.}
%
%
%
%
	
	
	\institute{
	\textsuperscript{1}Image Sciences Institute, University Medical Center Utrecht, the Netherlands \\
	\textsuperscript{2}Division of Image Processing, Leiden University Medical Center, the Netherlands}

	\maketitle
	
	\begin{abstract}
		In this work we propose a deep learning network for deformable image registration (DIRNet). 
		The DIRNet consists of a convolutional neural network (ConvNet) regressor, a spatial transformer, and a resampler. The ConvNet analyzes a pair of fixed and moving images and outputs parameters for the spatial transformer, which generates the displacement vector field that enables the resampler to warp the moving image to the fixed image. 
        The DIRNet is trained end-to-end by unsupervised optimization of a similarity metric between input image pairs. A trained DIRNet can be applied to perform registration on unseen image pairs in one pass, thus non-iteratively. 
        Evaluation was performed with registration of images of handwritten digits (MNIST) and cardiac cine MR scans (Sunnybrook Cardiac Data). The results demonstrate that registration with DIRNet is as accurate as a conventional deformable image registration method with substantially shorter execution times.

	\end{abstract}
	
	\section{Introduction}
	Image registration is a fundamental step in many clinical image analysis tasks. Traditionally, image registration is performed by exploiting intensity information between pairs of fixed and moving images. Since recently, deep learning approaches are used to aid image registration.
	Wu et al.~\cite{wu2016} used a convolutional stacked auto-encoder (CAE) to extract features from fixed and moving images that are subsequently used in conventional deformable image registration algorithms. However, the CAE is decoupled from the image registration task and hence, it does not necessarily extract the features most descriptive for image registration. The training of the CAE was unsupervised, but the registration task was not learned end-to-end. On the contrary, Miao et al.~\cite{miao2015} and Liao et al.~\cite{liao2016} have used deep learning to learn  rigid registration end-to-end. Miao et al.~\cite{miao2015} used a convolutional neural network (ConvNet) regressor to predict a transformation matrix for rigid registration of synthetic 2D to 3D images. Liao et al.~\cite{liao2016} used a ConvNet for intra-patient rigid registration of CT to cone-beam CT applied to either cardiac or abdominal images. This ConvNet learned to predict iterative updates of registration using reinforcement learning. Both registration methods were supervised: for training, transformation parameters were generated, which is task specific and highly challenging.

	Jaderberg et al.~\cite{jaderberg2015} introduced the spatial transformer network (STN) that can be used as a building block that aligns input images in a larger network that performs a particular task. By training the entire network end-to-end, the embedded STN deduces optimal alignment for solving that specific task. However, alignment is not guaranteed, and it is only performed when required for the task of the entire network. The STNs were used for affine transformations, as well as deformable transformations using thin-plate splines. However, an STN needs many labeled training examples, and to the best of our knowledge, have not yet been used in medical imaging.
	
	In this work, we present the deformable image registration network (DIRNet). The DIRNet takes pairs of fixed and moving images as inputs, and it outputs moving images warped to the fixed images. Training of the DIRNet is unsupervised. Unlike previous methods, the DIRNet is not trained with known registration transformations, but learns to register images by directly optimizing a similarity metric between the fixed and the moving image. Hence, similar to conventional intensity-based image registration, it directly learns the registration task end-to-end. In addition, a trained DIRNet is able to perform deformable image registration non-iteratively on unseen data. To the best of our knowledge, this is the first deep learning method for end-to-end unsupervised deformable image registration.

	\begin{figure}[t!]
		\center
		\includegraphics[width=\textwidth]{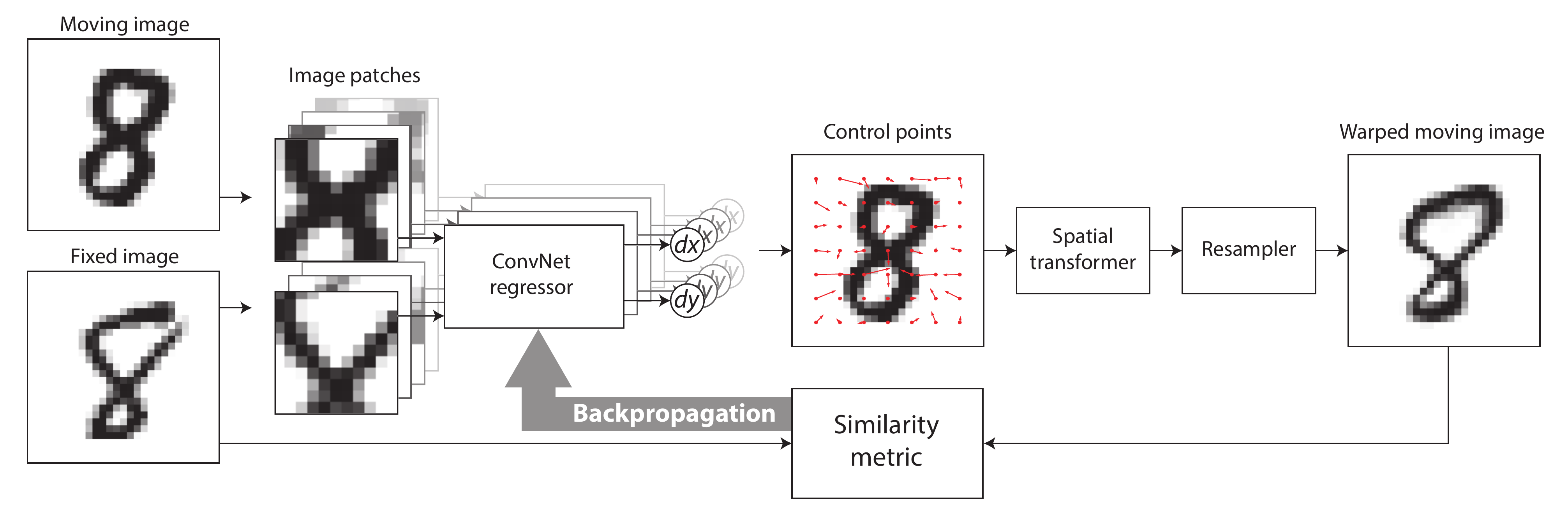}
		\caption{Schematics of the DIRNet with two input images from the MNIST data. The DIRNet takes one or more pairs of moving and fixed images as its inputs. The fully convolutional ConvNet regressor analyzes spatially corresponding image patches from the moving and fixed images and generates a grid of control points for a B-spline transformer. The B-spline transformer generates a full displacement vector field to warp a moving image to a fixed image. Training of the DIRNet is unsupervised and end-to-end by backpropagating an image similarity metric as a loss.}
		\label{fig:dirnet}
	\end{figure}
	
	\section{Method}
	\label{sec:method}
	The proposed DIRNet consists of a ConvNet regressor, a spatial transformer, and a resampler. The ConvNet regressor analyzes spatially corresponding patches from a pair of fixed and moving input images and outputs local deformation parameters for the spatial transformer. The spatial transformer generates a dense displacement vector field (DVF) that enables the resampler to warp the moving image to the fixed image. The DIRNet learns the registration task end-to-end by unsupervised training with an image similarity metric. Since the training phase involves simultaneous optimization of registration of many image pairs, the ConvNet implicitly learns a representation of the features in images that are important for predictions of local displacement. 
	Unlike regular image registration methods that typically perform iterative optimization for each image pair at hand, a trained DIRNet registers images in one pass.

	
	The ConvNet regressor expects concatenated pairs of moving and fixed images as its input, and applies four alternating layers of $3\times3$ convolutions with 0-padding and $2\times2$ downsampling layers. Downsampling reduces the number of the ConvNet parameters, but it is associated with translational invariance. We postulate that this effect should be minimal in a ConvNet used for image registration, thus we use average pooling which should retain the most information during downsampling. Subsequently, three $1\times1$ convolutional layers are applied to make the ConvNet regressor fully convolutional. Batch normalization~\cite{ioffe2015} is applied in every layer. Throughout the network exponential linear units~\cite{clevert2015} are used for activation, except for the final layer, which has a linear output. The number of kernels per layer can be of arbitrary size, but the number of kernels of the output layer is determined by the dimensionality of the input images (e.g. 2 kernels for 2D images that require 2D displacement). The fully convolutional design ensures analysis of separate but spatially corresponding fixed and moving image patch pairs with deformation parameters as outputs. The input image sizes and the number of downsampling layers jointly define the number of output parameters, i.e. the size and spacing of the control point grid. This way, for images of different sizes, similar grid spacing is ensured.
	Using the control point displacements, the spatial transformer generates a DVF used to warp the moving image to the fixed image. Like in \cite{jaderberg2015}, a thin-plate spline could be used as a spatial transformer, but due to its global support it is deemed less suitable for a patched-based approach. Therefore, we implemented a cubic B-spline~\cite{rueckert1999} transformer which has local support.

	The DIRNet is trained by optimizing an image similarity metric (i.e. by backpropagating dissimilarity) between pairs of moving and fixed images from a training set using mini-batch stochastic gradient descent (Adam~\cite{kingma2014}). Any similarity metric used in conventional image registration could be used. In this work normalized cross correlation is employed.
    After training, the DIRNet can be applied for registration of unseen images.

    \section{Data}
	The DIRNet was evaluated with handwritten digits from the MNIST database~\cite{lecun1998mnist} and clinical MRI scans from the Sunnybrook Cardiac Data (SCD)~\cite{Radau2009}.
	
	The MNIST database contains $28\times28$ pixel grayscale images of handwritten digits that were centered by computing the center of mass of the pixels.
	The test images (10,000 digits) were kept separate from the training images (60,000 digits). One sixth of the training data was used for validation to monitor overfitting during training.
	
	The SCD contains 45 cardiac cine MRI scans that were acquired on a single MRI-scanner. The scans consist of short-axis cardiac image slices each containing 20 timepoints that encompass the entire cardiac cycle. Slice thickness and spacing is 8\,mm, and slice dimensions are $256\times256$ with a pixel size of $1.28$\,mm$\times1.28$\,mm. The SCD is equally divided in 15 training scans (183 slices), 15 validation scans (168 slices), and 15 test scans (176 slices). An expert annotated the left ventricle myocardium at end-diastolic (ED) and end-systolic (ES) time points following the annotation protocol of the SCD. Annotations were made in the test scans and only used for final quantitative evaluation. In total, 129 image slices were annotated, i.e. 258 annotated timepoints.

	\section{Experiments and Results}
	DIRNet was implemented with Theano\footnote{http://deeplearning.net/software/theano/ \textit{(version 0.8.2)}} and Lasagne\footnote{https://lasagne.readthedocs.io/en/latest/ \textit{(version 0.2.dev1)}}, and conventional registration was performed with SimpleElastix~\cite{Marstal2016}.



	\subsection{Registration of handwritten digits}
\begin{figure}[t!]
		\centering
		\includegraphics[width = .4\textwidth, trim = 0 0 0 0, clip]{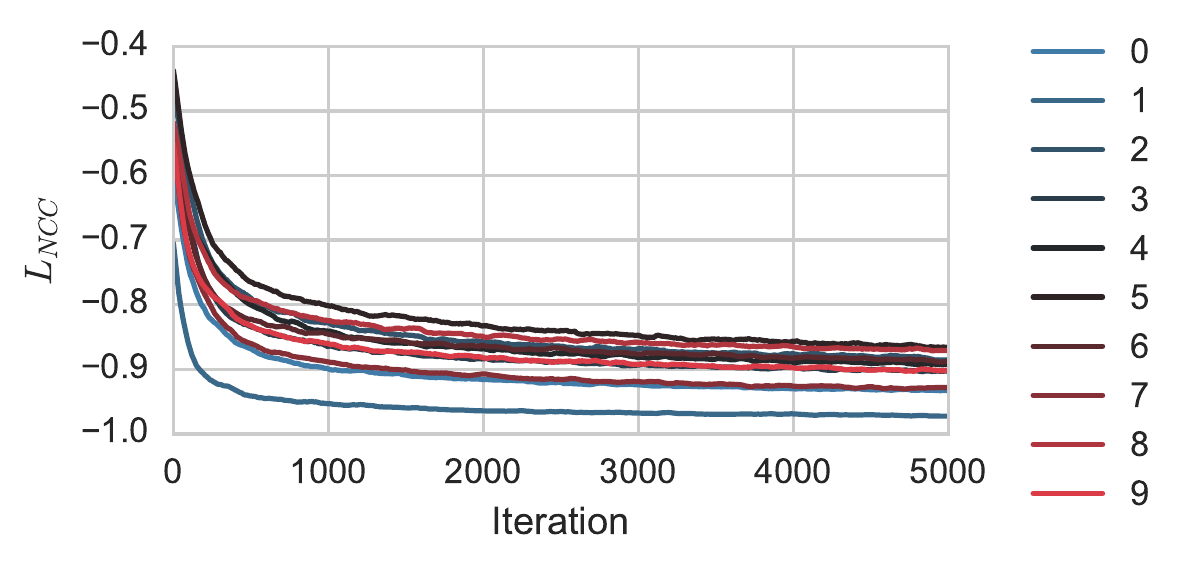}
		\includegraphics[width = 0.05\textwidth, trim = 20 0 10 0, clip]{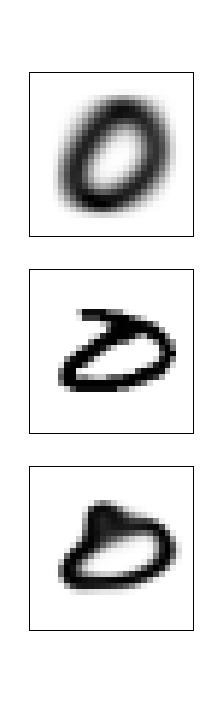}
		\includegraphics[width = 0.05\textwidth, trim = 20 0 10 0, clip]{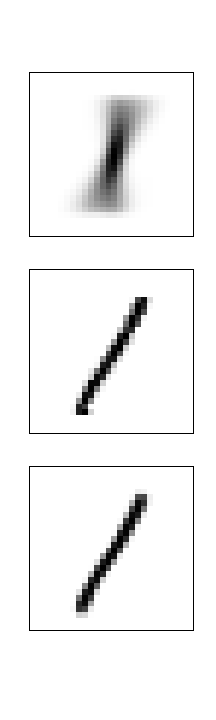}
		\includegraphics[width = 0.05\textwidth, trim = 20 0 10 0, clip]{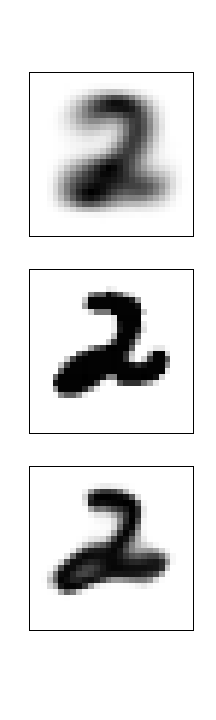}
		\includegraphics[width = 0.05\textwidth, trim = 20 0 10 0, clip]{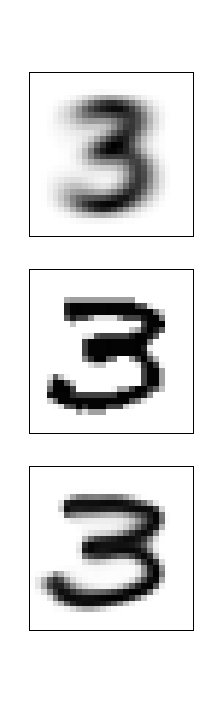}
		\includegraphics[width = 0.05\textwidth, trim = 20 0 10 0, clip]{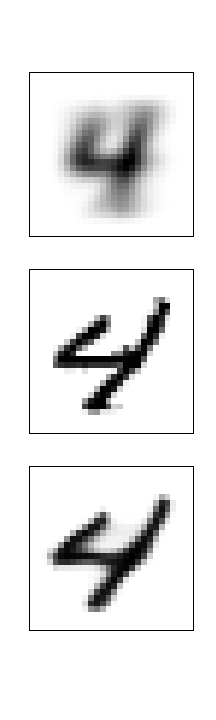}
		\includegraphics[width = 0.05\textwidth, trim = 20 0 10 0, clip]{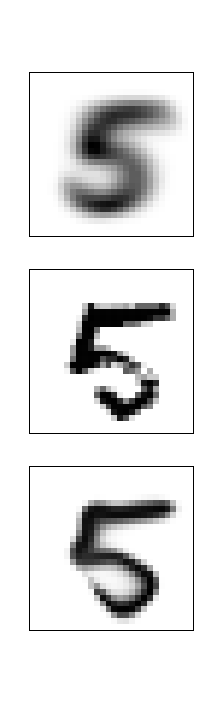}
		\includegraphics[width = 0.05\textwidth, trim = 20 0 10 0, clip]{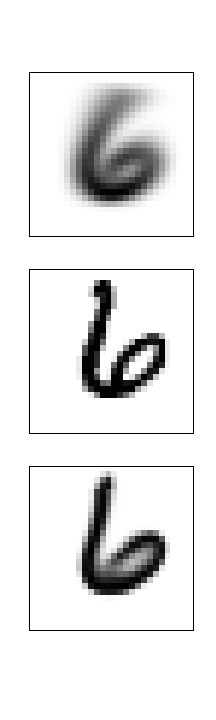}
		\includegraphics[width = 0.05\textwidth, trim = 20 0 10 0, clip]{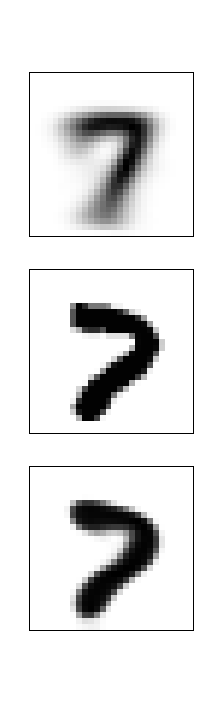}
		\includegraphics[width = 0.05\textwidth, trim = 20 0 10 0, clip]{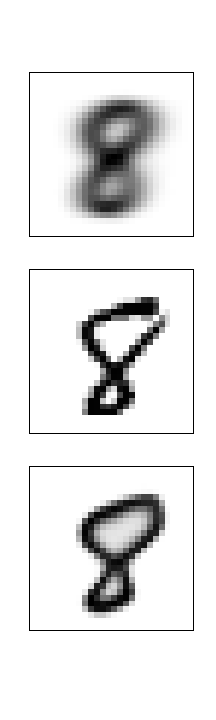}
		\includegraphics[width = 0.05\textwidth, trim = 20 0 10 0, clip]{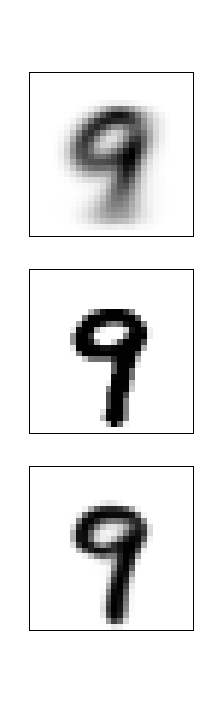}
		\vspace{-.3cm}
		\caption{Left: Learning curves showing the normalized cross correlation loss ($L_\mathrm{NCC}$) on the validation set of DIRNets trained in 5,000 iterations for registration of MNIST digits. Right: Registration results of the trained DIRNets on a separate test set. The top row shows an average of all moving images per class (about 1,000 digits), the middle row shows one randomly chosen fixed image per class, and the bottom row shows an average of the registration results of independent registrations of the moving images to the chosen fixed image.
		}
		\label{fig:mnist:results}
	\end{figure}

	Seperate DIRNet instances were trained for image registration of a specific class: one for each digit. The DIRNets were designed with 16 kernels per convolution layer, the third and fourth downsampling layers were removed. This resulted in a  control point grid of $7\times7$ (grid spacing of 4 pixels).
	Each DIRNet was trained separately with random combinations of digits from its class with mini-batches of 32 random fixed and moving image pairs in 5,000 iterations (i.e. backpropagations). See Figure~\ref{fig:mnist:results} (left) for the learning curves.
	
	Registration performance of the trained DIRNets was qualitatively assessed on the test data. For each digit, one sample was randomly chosen to be the fixed image. Thereafter, all remaining digits (approximately 1,000 per class) were registered to the corresponding fixed image. Figure~\ref{fig:mnist:results} (right) shows the registration results.

	\subsection{Registration of cardiac MRI}
	The DIRNet was trained by randomly selecting pairs of fixed and moving image slices from cardiac cine MRI scans (4D data). The pairs of fixed and moving images were anatomically corresponding slices from the same 4D scan of a single patient but acquired at different time points in the cardiac cycle. This resulted in 69,540 image pairs for training, and 63,840 pairs for validation.
	
	

	A baseline DIRNet, as described in Section~\ref{sec:method}, was designed with 16 kernels per convolution layer. This resulted in a grid of $16\times16$ control points, i.e. a grid spacing of 16 pixels (20.48\,mm). To evaluate effect of various DIRNet parameters, additional experiments were performed. First, to evaluate the effect of the downsampling method, DIRNet-A1 was designed with max-pooling layers, and DIRNet-A2 was designed with $2\times2$ strided convolutions. Second, to evaluate the effect of the spatial transformer, DIRNet-B1 was designed with a quadratic B-spline transformer, and DIRNet-B2 with a thin-plate spline transformer. Finally, to show the effect of the size of the receptive field (i.e. patch size), DIRNet-C1 was designed with overlapping patches such that they coincided with the capture range of the B-spline control points. This was achieved by adding an extra $3\times3$ convolutional layer before and after the final pooling layer. In addition, DIRNet-C2 analyzed full image slices for each control point by replacing the $1\times1$ convolution layers with a $3\times3$ convolution layer, followed by a downsampling layer, two fully connected layers of 1,024 nodes, and a final output layer of $16\times16$ 2D control points.
	
\begin{figure}[t!]
		\centering
    		\includegraphics[width = .8\textwidth, trim = 0 0 0 0, clip]{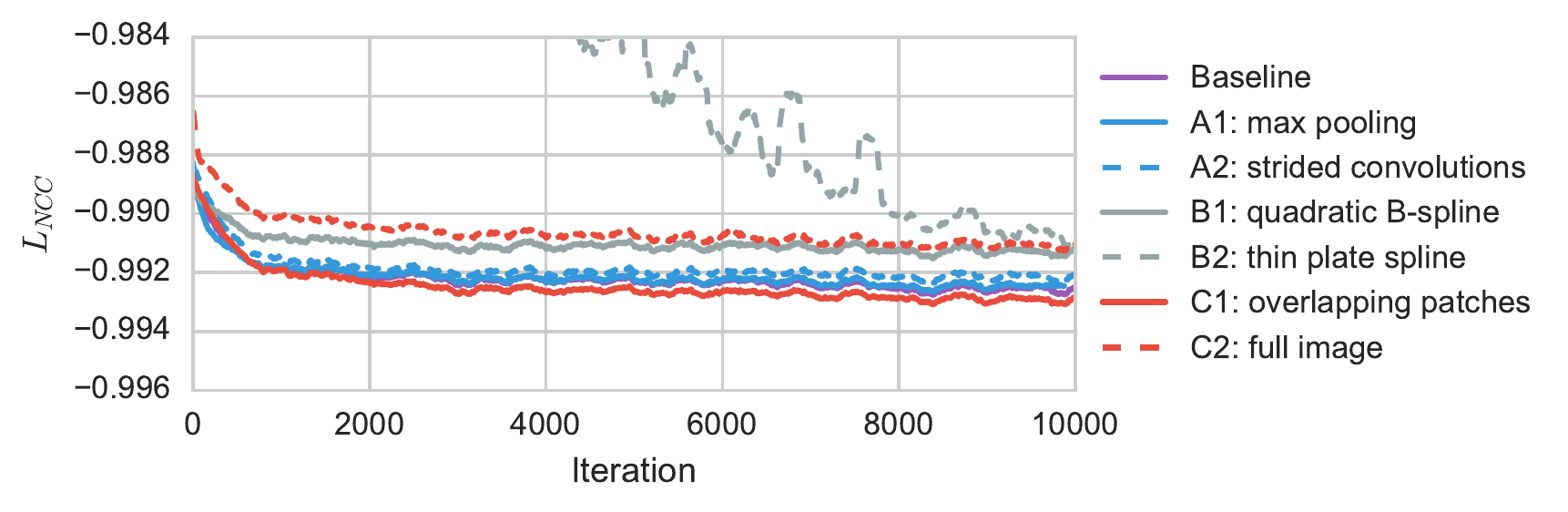}
		\vspace{-.4cm}
		\caption{Validation loss over 10,000 iteration for the baseline DIRNet, DIRNets with different downsampling techniques (A1, A2), DIRNets with different spatial transformers (B1, B2), and DIRNets with different receptive fields (C1, C2).}
		\label{fig:cardiacmr:learning_curves}
	\end{figure}

	Each DIRNet was trained until convergence in mini-batches of 32 image pairs in at least 10,000 iterations. The training loss closely followed the validation loss in each experiment, and no signs of overfitting were apparent. Figure~\ref{fig:cardiacmr:learning_curves} shows the validation loss of 10,000 iterations during training for all experiments. The DIRNets converged quickly in each experiment, except DIRNet-B2, where convergence was reached after approximately 30,000 iterations, but with a loss greater than the baseline DIRNet. The final loss was lowest for DIRNet-C1.

	\begin{table}[t!]
        \centering
        \caption{Quantitative cardiac MRI registration results by comparing reference annotations in fixed images and warped annotations of the moving to the fixed images. The table lists mean and standard deviation for the Dice score, 95\textsuperscript{th} percentiles of the surface distance (95\textsuperscript{th}SD), and mean absolute surface distance (MAD). The rows show results before registration, with conventional iterative image registration using SimpleElastix, and registration using the DIRNet. The best obtained results are shown in bold.}
        \label{tab:quantitative_results}
            \begin{tabular}{lcC{1.8cm}C{1.8cm}C{1.8cm}}
                &           & Dice & 95\textsuperscript{th}SD & MAD \\
                \hline
                \multicolumn{2}{l}{Before} & $0.62\pm0.15$ & $7.79\pm2.92$ & $2.89\pm1.07$\\
                \hline
                \multicolumn{2}{l}{SimpleElastix}  &$0.79\pm0.08$ &$5.09\pm2.36$ &$1.91\pm0.94$ \\
                \hline
                DIRNet& BL  &$0.79\pm0.08$ &$5.20\pm2.30$ &$1.92\pm0.89$ \\
                & A1        &$0.78\pm0.08$ &$5.26\pm2.16$ &$1.95\pm0.85$ \\
                & A2        &$0.78\pm0.08$ &$5.30\pm2.28$ &$1.97\pm0.87$ \\
                & B1        &$0.72\pm0.11$ &$6.41\pm2.61$ &$2.40\pm0.96$ \\
                & B2        &$0.78\pm0.09$ &$5.48\pm2.36$ &$2.01\pm0.89$ \\
                & C1        &$\bm{0.80\pm0.08}$ &$\bm{5.03\pm2.30}$ &$\bm{1.83\pm0.89}$. \\
                & C2        &$0.76\pm0.09$ &$5.55\pm2.24$ &$2.10\pm0.90$\\
         
            \end{tabular}
    \end{table}
	
\begin{figure}[t!]
		\center
		\begin{tabular}{cccc}
		    Fixed image & Moving image & DIRNet  & SimpleElastix \\
			\includegraphics[width = .24\textwidth, trim = 0 0 0 0, clip] {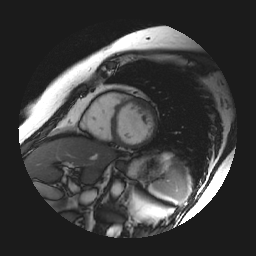}
			&
			\includegraphics[width = .24\textwidth, trim = 0 0 0 0, clip] {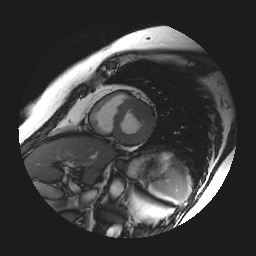}
			&
			
			\includegraphics[width = .24\textwidth, trim = 0 0 0 0, clip] {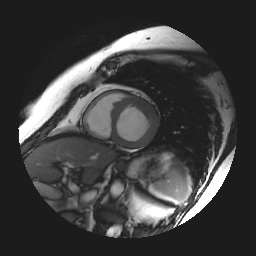}
			&
			\includegraphics[width = .24\textwidth, trim = 0 0 0 0, clip] {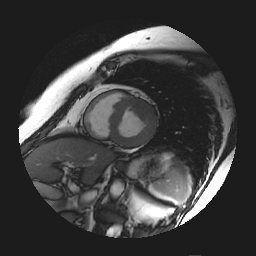}
			\\
			&
			\includegraphics[width = .24\textwidth, trim = 0 0 0 0, clip] {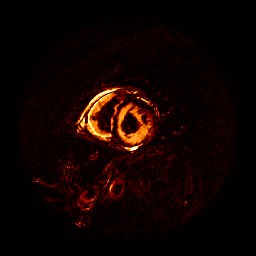}
			&
			\includegraphics[width = .24\textwidth, trim = 0 0 0 0, clip] {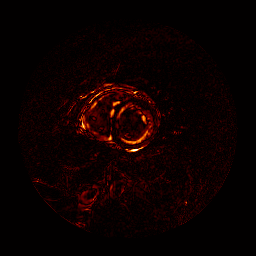}
			&
			\includegraphics[width = .24\textwidth, trim = 0 0 0 0, clip] {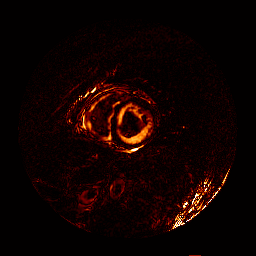}
		\end{tabular}
		\vspace{-.2cm}
		\caption{Top, from left to right: The fixed (ED), the moving (ES), the DIRNet-C1 warped, and the SimpleElastix warped images. Bottom: Heatmaps showing absolute difference images between the fixed image and (from left to right) the original, the DIRNet warped, and the SimpleElastix warped moving images.}
		\label{fig:cardiacmr:example}
		
	\end{figure}

	Quantitative evaluation was performed on the test set by registering image slices at ED to ES, and vice versa, which resulted in 258 independent registration experiments. The obtained transformation parameters were used to warp the left ventricle annotations of the moving image to the fixed image. The transformed annotations were compared with the reference annotations in the fixed images. The results are listed in Table~\ref{tab:quantitative_results}. The best registration results were obtained with the DIRNet-C1 on an NVIDIA Titan X Maxwell GPU in $0.049\pm0.0035$\,s.	
	For comparison, the table also lists conventional iterative intensity-based registrations (SimpleElastix), with parameters specifically tuned for this task. SimpleElastix used a similar grid spacing as the DIRNet but with a multiresolution approach, downsampling first with a factor of 2 and thereafter using the original resolution. Updating in 100 iterations per resolution was sufficient for convergence with a reasonable time span. Experiments were performed with an Intel Xeon 1620-v3 3.5\,GHz CPU using 8 threads in $0.51\pm0.070$\,s (10 times slower than the DIRNet). Figure~\ref{fig:cardiacmr:example} shows registration results for a randomly chosen image pair.


	\section{Discussion and Conclusion}
	A deep learning method for unsupervised end-to-end learning of deformable image registration has been presented. The method has been evaluated with registration of images with handwritten digits and image slices from cine cardiac MRI scans.
	The presented DIRNet achieves a performance that is as accurate as a conventional deformable image registration method with substantially shorter execution times. 
	The method does not require training data, which is often difficult to obtain for medical images.
	To the best of our knowledge this is the first deep learning method that uses unsupervised end-to-end training for deformable image registration.
	
	Even though registration of images with handwritten digits is an easy  task, the performed experiments demonstrate that a single DIRNet architecture can be used to perform registration in different image domains given domain specific training. It would be interesting to further investigate whether a single DIRNet instance can be trained for registration across different image domains.
	
	
	Registration of slices from cardiac cine MRI scans was quantitatively evaluated between the ES and ED timepoints, so at maximum cardiac compression and maximum dilation. Even though conventional registration method (SimpleElastix) was specifically tuned for this task, and the DIRNet was trained by registration of slices from any timepoint of the cardiac cycle, the results demonstrate that the results of the DIRNet instances were  either comparable or slightly outperformed the conventional approach (DirNet-C1 vs. SimpleElastix: p $\ll0.01$).

	The data used in this work did not require rough pre-alignment of images, by e.g. affine registration. However, to extend the applicability of the proposed method in future work, performing affine registration will be investigated. Furthermore, proposed method is designed for registration of 2D images. In future work the method will be extended to perform registration of 3D images. Moreover, experiments were performed using only normalized cross correlation as an image similarity metric, but any differentiable metric could be used.

	To conclude, the DIRNet is able to learn image registration tasks in an unsupervised end-to-end fashion using an image similarity metric for optimization. 
	Image registration is performed in one pass, thus non-iteratively. 
	The results demonstrate that the network achieves a performance that is as accurate as a conventional deformable image registration method within substantially shorter execution times.
	
	\section*{Acknowledgment}

 	This study was funded by the Netherlands Organization for Scientific Research (NWO); Project 12726.
	
    \bibliography{bibliography}{}
	\bibliographystyle{splncs03}

\end{document}